# CALIBRATION OF QUASI-ISOTROPIC PARALLEL KINEMATIC MACHINES: ORTHOGLIDE

Anatoly Pashkevich, Roman Gomolitsky
*Robotic Laboratory, Department of Control Systems*
*Belarusian State University of Informatics and Radioelectronics*
*6 P.Brovka St., Minsk 220027, Belarus*
*pap@bsuir.unibel.by*

Philippe Wenger, Damien Chablat
*Institut de Recherche en Communications et Cybernétique de Nantes*
*1, rue de la Noe B.P. 6597, 44321 Nantes Cedex 3, France*
*{ Philippe.Wenger, Damien.Chablat}@irccyn.ec-nantes.fr*

Abstract:   The paper proposes a novel approach for the geometrical model calibration of quasi-isotropic parallel kinematic mechanisms of the Orthoglide family. It is based on the observations of the manipulator leg parallelism during motions between the specific test postures and employs a low-cost measuring system composed of standard comparator indicators attached to the universal magnetic stands. They are sequentially used for measuring the deviation of the relevant leg location while the manipulator moves the TCP along the Cartesian axes. Using the measured differences, the developed algorithm estimates the joint offsets and the leg lengths that are treated as the most essential parameters. Validity of the proposed calibration technique is confirmed by the experimental results.

## 1 INTRODUCTION

Parallel kinematic machines (PKM) are commonly claimed as appealing solutions in many industrial applications due to their inherent structural rigidity, good payload-to-weight ratio, high dynamic capacities and high accuracy (Tlusty et al., 1999; Merlet, 2000; Wenger et al., 2001). However, while the PKM usually exhibit a much better repeatability compared to serial mechanisms, they may not necessarily posses a better accuracy, which is limited by manufacturing/assembling errors in numerous links and passive joints (Wang and Masory, 1993). Thus, the PKM accuracy highly relies on the accurate kinematic model, which must be carefully tuned for each manipulator separately.

Similar to the serial manipulators, the PKM calibration techniques are based on the minimization of a parameter-dependent error function, which incorporates residuals of the kinematic equations. For the parallel manipulators, the inverse kinematic equations are considered computationally more efficient (contrary to the direct kinematics, which is usually analytically unsolvable for the PKM) (Innocenti, 1995; Iurascu & Park, 2003; Jeong et al., 2004, Huang et al., 2005). But the main difficulty with this technique is the full-pose measurement requirement, which is very hard to implement accurately. Hence, a number of studies have been directed at using the subset of the pose measurement data, which however creates another problem, the identifiability of the model parameters.

Popular approaches in the parallel robot calibration deal with one-dimensional pose errors using a double-ball-bar system or other measuring devices, as well as imposing mechanical constraints on some elements of the manipulator (Daney, 2003). However, in spite of hypothetical simplicity, it is hard to implement in practice since an accurate extra mechanism is required to impose these constraints. Additionally, such methods reduce the workspace size and consequently the identification efficiency.

Another category of the methods, the self- or autonomous calibration, is implemented by minimizing the residuals between the computed and measured values of the active and/or redundant joint sensors. Adding extra sensors at the usually unmeasured joints is very attractive from computational point of view, since it allows getting the data in the whole workspace and potentially reduces impact of the measurement noise. However, only a partial set of the parameters may be identified in this way, since the internal sensing is unable to provide sufficient information for the robot end-effector absolute location.

More recently, several hybrid calibration methods were proposed that utilize intrinsic properties of a particular parallel machine allowing extracting the full set of the model parameters (or the most essential of them) from a minimum set of measurements. It worth mentioning an innovative approach developed by Renaud et al. (2004, 2005) who applied the vision-based measurement system for the PKM calibration from the leg observations. In this technique, the source data are extracted from the leg images, without any strict assumptions on the end-effector poses. The only assumption is related to the manipulator architecture (the mechanism is actuated by linear drives located on the base). However, current accuracy of the camera-based measurements is not high enough yet to apply this method in industrial environment.

This paper extends our previous research (Pashkevich et al., 2006) and focuses on the calibration of the Orthoglide-type mechanisms, which is also actuated by linear drives located on the manipulator base and admits technique of Renaud et al. (2004, 2005). But, in contrast to the known works, our approach assumes that the leg location is observed for specific manipulator postures, when the tool-center-point moves along the Cartesian axes. For these postures and for the nominal Orthoglide geometry, the legs are strictly parallel to the corresponding Cartesian planes. So, the deviation of the manipulator parameters influences on the leg parallelism that gives the source data for the parameter identification. The main advantage of this approach is the simplicity and low cost of the measuring system that can avoid using computer vision and is composed of standard comparator indicators attached to the universal magnetic stands.

The remainder of the paper is organized as follows. Section 2 describes the manipulator geometry, its inverse and direct kinematics, and also contains the sensitivity analysis of the leg parallelism at the examined postures with respect to the geometrical parameters. Section 3 focuses on the parameter identification, with particular emphasis on the calibration accuracy under the measurement noise. Section 4 contains experimental results that validate the proposed technique, while Section 5 summarizes the main contributions.

## 2 ORTHOGLIDE MECHANISM

### 2.1 Manipulator architecture

The Orthoglide is a three degrees-of-freedom parallel manipulator actuated by linear drives with mutually orthogonal axes. Its kinematic architecture is presented in Figure 1 and includes three identical parallel chains, which will be further referred as "legs". Each manipulator leg is formally described as $PRP_aR$ - chain, where $P$, $R$ and $P_a$ denote the prismatic, revolute, and parallelogram joints respectively (Figure 2). The output machinery (with a tool mounting flange) is connected to the legs in such a manner that the tool moves in the Cartesian space with fixed orientation (translational motions).

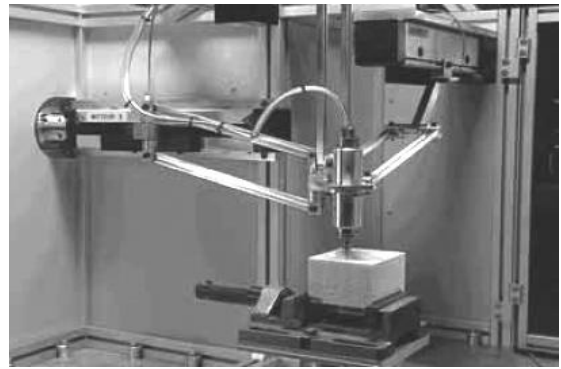

Figure 1: The Orthoglide kinematic architecture.
(© CNRS Photothèque / CARLSON Leif)

The Orthoglide workspace has a regular, quasi-cubic shape. The input/output equations are simple and the velocity transmission factors are equal to one along the x, y and z direction at the isotropic configuration, like in a conventional serial PPP machine (Wenger et al., 2000; Chablat and Wenger, 2003). The latter is an essential advantage of the Orthoglide architecture, which also allows referring it as the "quasi-isotropic" kinematic machine.

Another specific feature of the Orthoglide mechanism, which will be further used for the calibration, is displayed during the end-effector motions along the Cartesian axes. For example, for the x-axis motion in the Cartesian space, the sides of the x-leg parallelogram must also retain strictly parallel to the x-axis. Hence, the observed deviation of the mentioned parallelism may be used as the data source for the calibration algorithms.

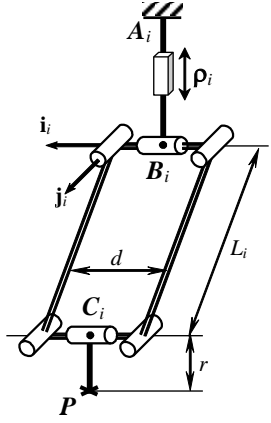

Figure 2: Kinematics of the Orthoglide leg.

For a small-scale Orthoglide prototype used for the calibration experiments, the workspace size is approximately equal to 200×200×200 mm$^3$ with the velocity transmission factors bounded between 1/2 and 2 (Chablat & Wenger, 2003). The legs nominal geometry is defined by the following parameters: $L_i$ = 310.25 mm, $d$ = 80 mm, $r$ = 31 mm where $L_i$, $d$ are the parallelogram length and width, and $r$ is the distance between the points $C_i$ and the tool centre point $P$ (see Figure 2).

## 2.2 Modelling assumptions

Following previous studies on the PKM accuracy (Wang & Massory, 1993; Renaud et al., 2004), the influence of the joint defects is assumed negligible compared to the encoder offsets and the link length deviations. This validates the following modelling assumptions:

(i) the manipulator parts are supposed to be rigid bodies connected by perfect joints;
(ii) the manipulator legs (composed of a prismatic joint, a parallelogram, and two revolute joints) generate a four degrees-of-freedom motions;
(iii) the articulated parallelograms are assumed to be perfect but non-identical;
(iv) the linear actuator axes are mutually orthogonal and intersected in a single point to insure a translational movement of the end-effector;
(v) the actuator encoders are perfect but located with some errors (offsets).

Using these assumptions, there will be derived new calibration equations based on the observation of the parallel motions of the manipulator legs.

## 2.3 Basic equations

Since the kinematic parallelograms are admitted to be non-identical, the kinematic model developed in in our previous papers (Pashkevich et al., 2005, 2006) should be extended to describe the manipulator with different length of the legs.

Under the adopted assumptions, similar to the equal-leg case, the articulated parallelograms may be replaced by the kinematically equivalent bar links. Besides, a simple transformation of the Cartesian coordinates (shift by the vector $(r, r, r)^T$, see Figure 2) allows to eliminate the tool offset. Hence, the Orthoglide geometry can be described by a simplified model, which consists of three rigid links connected by spherical joints to the tool centre point (TCP) at one side and to the allied prismatic joints at another side. Corresponding formal definition of each leg can be presented as PSS, where P and S denote the actuated prismatic joint and the passive spherical joint respectively.

Thus, if the origin of a reference frame is located at the intersection of the prismatic joint axes and the $x$, $y$, $z$-axes are directed along them, the manipulator kinematics may be described by the following equations

$$\mathbf{p} = \begin{bmatrix} (\rho_x + \Delta\rho_x) + \cos\theta_x \cos\beta_x L_x + e \\ \sin\theta_x \cos\beta_x L_x \\ -\sin\beta_x L_x \end{bmatrix}; \quad (1a)$$

$$\mathbf{p} = \begin{bmatrix} -\sin\beta_y L_y \\ (\rho_y + \Delta\rho_y) + \cos\theta_y \cos\beta_y L_y + e \\ \sin\theta_y \cos\beta_y L_y \end{bmatrix}; \quad (1b)$$

$$\mathbf{p} = \begin{bmatrix} \sin\theta_z \cos\beta_z L_z \\ -\sin\beta_z L_z \\ (\rho_z + \Delta\rho_z) + \cos\theta_z \cos\beta_z L_z + e \end{bmatrix}, \quad (1c)$$

where $\mathbf{p} = (p_x, p_y, p_z)^T$ is the output vector of the TCP position, $\boldsymbol{\rho} = (\rho_x, \rho_y, \rho_z)^T$ is the input vector of the prismatic joints variables, $\Delta\boldsymbol{\rho} = (\Delta\rho_x, \Delta\rho_y, \Delta\rho_z)^T$ is the encoder offset vector, $\theta_i$, $\beta_i$, $i \in \{x, y, z\}$ are the parallelogram orientation angles (internal variables), and $L_i$ are the length of the corresponding leg.

After elimination of the internal variables $\theta_i$, $\beta_i$, the kinematic model (1) can be reduced to three equations

$$(p_i - (\rho_i + \Delta\rho_i))^2 + p_j^2 + p_k^2 = L_i^2, \quad (2)$$

which includes components of the input and output vectors **p** and **ρ** only. Here, the subscripts $i, j, k \in \{x, y, z\}$, $i \neq j \neq k$ are used in all combinations, and the joint variables $\rho_i$ are obeyed the prescribed limits $\rho_{min} < \rho_i < \rho_{max}$ defined in the control software (for the Orthoglide prototype, $\rho_{min}$ = -100 mm and $\rho_{max}$ = +60 mm).

It should be noted that, for the case $\Delta\rho_x = \Delta\rho_y = \Delta\rho_z = 0$ and $L_x = L_y = L_z = L$, the nominal ''mechanical-zero'' posture of the manipulator corresponds to the Cartesian coordinates $\mathbf{p}_0 = (0, 0, 0)^T$ and to the joints variables $\boldsymbol{\rho}_0 = (L, L, L)$. Moreover, in such posture, the *x*-, *y*- and *z*-legs are oriented strictly parallel to the corresponding Cartesian axes. But the joint offsets and the leg length differences cause the deviation of the "zero" TCP location and corresponding deviation of the leg parallelism, which may be measured and used for the calibration.

Hence, six parameters ($\Delta\rho_x, \Delta\rho_y, \Delta\rho_z, L_x, L_y, L_z$) define the manipulator geometry and are in the focus of the proposed calibration technique.

## 2.4 Inverse and direct kinematics

The *inverse kinematic* relations are derived from the equations (2) in a straightforward way and only slightly differ from the "nominal" case:

$$\rho_i = p_i + s_i \sqrt{L_i^2 - p_j^2 - p_k^2} - \Delta\rho_i, \quad (3)$$

where $s_x, s_y, s_z \in \{\pm 1\}$ are the configuration indices defined for the "nominal" geometry as the signs of $\rho_x - p_x$, $\rho_y - p_y$, $\rho_z - p_z$, respectively. It is obvious that expressions (3) give eight different solutions, however the Orthoglide prototype assembling mode and the joint limits reduce this set to a single case corresponding to the $s_x = s_y = s_z = 1$.

For the direct kinematics, equations (2) can be subtracted pair-to-pair that gives linear relations between the unknowns $p_x$, $p_y$, $p_z$, which may be expressed in the parametric form as

$$p_i = \frac{\rho_i + \Delta\rho_i}{2} + \frac{t}{\rho_i + \Delta\rho_i} - \frac{L_i^2}{2(\rho_i + \Delta\rho_i)}, \quad (4)$$

where $t$ is an auxiliary scalar variable. This reduces the direct kinematics to the solution of a quadratic equation $At^2 + Bt + C = 0$ with the coefficients

$$A = \sum_{i \neq j} (\rho_i + \Delta\rho_i)^2 (\rho_j + \Delta\rho_j)^2;$$

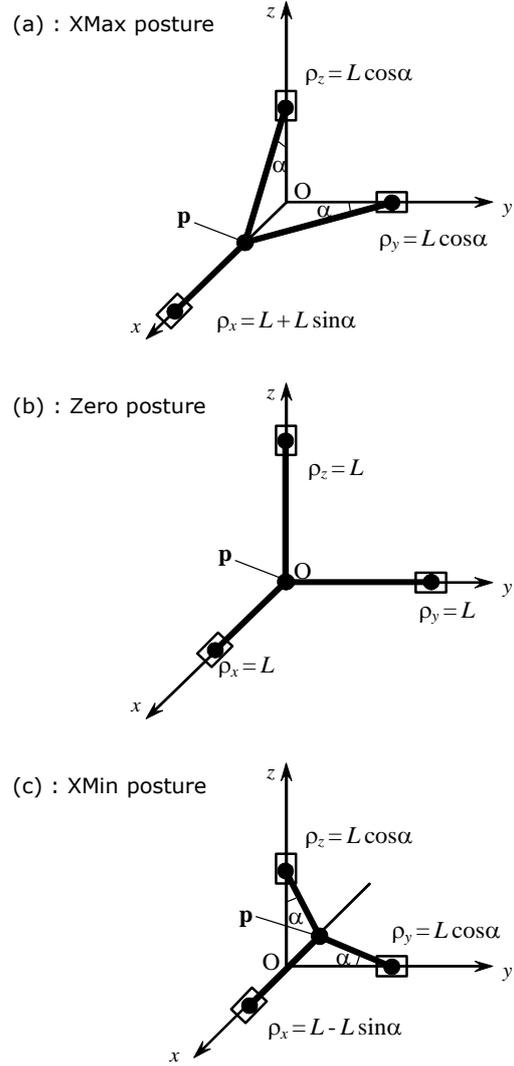

Figure 3: Specific postures of the Orthoglide (for the x-leg motion along the Cartesian axis X).

$$B = \prod_i (\rho_i + \Delta\rho_i)^2 - \sum_{i \neq j \neq k} L_i^2 (\rho_j + \Delta\rho_j)^2 (\rho_k + \Delta\rho_k)^2;$$

$$C = \prod_i (\rho_i + \Delta\rho_i)^2 \cdot \left( \sum_i (\rho_i + \Delta\rho_i)^2 / 4 - \sum_i L_i^2 / 2 \right) + \sum_{i \neq j \neq k} L_i^4 (\rho_j + \Delta\rho_j)^2 (\rho_k + \Delta\rho_k)^2 / 4$$

where $i, j, k \in \{x, y, z\}$. From two possible solutions that gives the quadratic formula, the Orthoglide prototype (see Figure 1) admit a single one $t = (-B + \sqrt{B^2 - 4AC})/2A$ corresponding to the manipulator assembling mode.

## 2.5 Differential relations

To obtain the calibration equations, first let us derive the differential relations for the TCP deviation for three types of the Orthoglide postures:
(i) "*maximum displacement*" postures for the directions x, y, z (Figure 3a);
(ii) "*mechanical zero*" or the isotropic posture (Figure 3b);
(iii) "*minimum displacement*" postures for the directions x, y, z (Figure 3c);

These postures are of particular interest for the calibration since, in the "nominal" case, a corresponding leg is parallel to the relevant pair of the Cartesian planes.

The manipulator Jacobian with respect to the parameters $\Delta\rho = (\Delta\rho_x, \Delta\rho_y, \Delta\rho_z)$ and $L = (L_x, L_y, L_z)$ can be derived by straightforward differentiating of the kinematic equations (2), which yields

$$\begin{bmatrix} p_x-\rho_x & p_y & p_z \\ p_x & p_y-\rho_y & p_z \\ p_x & p_y & p_z-\rho_z \end{bmatrix} \cdot \frac{\partial \mathbf{p}}{\partial \mathbf{\rho}} = \begin{bmatrix} p_x-\rho_x & 0 & 0 \\ 0 & p_y-\rho_y & 0 \\ 0 & 0 & p_z-\rho_z \end{bmatrix}$$

$$\begin{bmatrix} p_x-\rho_x & p_y & p_z \\ p_x & p_y-\rho_y & p_z \\ p_x & p_y & p_z-\rho_z \end{bmatrix} \cdot \frac{\partial \mathbf{p}}{\partial \mathbf{L}} = \begin{bmatrix} L_x & 0 & 0 \\ 0 & L_y & 0 \\ 0 & 0 & L_z \end{bmatrix}.$$

Thus, after the matrix inversions and multiplications, the desired Jacobian can be written as

$$\mathbf{J}(\mathbf{p},\mathbf{\rho}) = \left[ \mathbf{J}_\rho(\mathbf{p},\mathbf{\rho}); \ \mathbf{J}_L(\mathbf{p},\mathbf{\rho}) \right], \quad (5)$$

where

$$\mathbf{J}_\rho(.) = \begin{bmatrix} 1 & \dfrac{p_y}{p_x-\rho_x} & \dfrac{p_z}{p_x-\rho_x} \\ \dfrac{p_x}{p_y-\rho_y} & 1 & \dfrac{p_z}{p_y-\rho_y} \\ \dfrac{p_x}{p_z-\rho_z} & \dfrac{p_y}{p_z-\rho_z} & 1 \end{bmatrix}^{-1}$$

$$\mathbf{J}_L(.) = \begin{bmatrix} \dfrac{p_x-\rho_x}{L_x} & \dfrac{p_y}{L_x} & \dfrac{p_z}{L_x} \\ \dfrac{p_x}{L_y} & \dfrac{p_y-\rho_y}{L_y} & \dfrac{p_z}{L_y} \\ \dfrac{p_x}{L_z} & \dfrac{p_y}{L_z} & \dfrac{p_z-\rho_z}{L_z} \end{bmatrix}^{-1}$$

It should be noted that, for the computing convenience, the above expression includes both the Cartesian coordinates $p_x$, $p_y$, $p_z$ and the joint coordinates $\rho_x$, $\rho_y$, $\rho_z$, but only one of these sets may be treated as an independent taking into account the inverse/direct kinematic relations.

For the "*Zero*" posture, the differential relations are derived in the neighbourhood of the point $\{p_0 = (0, 0, 0) ; \ \rho_0 = (L, L, L)\}$, which after substitution to (5) gives the Jacobian matrix

$$\mathbf{J}_0 = \begin{bmatrix} 1 & 0 & 0 & -1 & 0 & 0 \\ 0 & 1 & 0 & 0 & -1 & 0 \\ 0 & 0 & 1 & 0 & 0 & -1 \end{bmatrix}. \quad (6)$$

Hence, in this case, the TCP displacement is related to the joint offsets and the leg legs variations $\Delta L_i$ by trivial equations

$$\Delta p_i = \Delta \rho_i - \Delta L_i; \ i \in \{x, y, z\}. \quad (7)$$

For the "*XMax*" posture, the Jacobian is computed in the neighbourhood of the point $\{ \mathbf{p} = (LS_\alpha, 0, 0) ; \ \mathbf{\rho} = (L+LS_\alpha, LC_\alpha, LC_\alpha) \}$, where $\alpha$ is the angle between the y-, z-legs and the X-axes: $\alpha = a\sin(\rho_{max}/L)$; $S_\alpha = \sin(\alpha)$, $C_\alpha = \cos(\alpha)$. This gives the Jacobian

$$\mathbf{J}_x^+ = \begin{bmatrix} 1 & 0 & 0 & -1 & 0 & 0 \\ T_\alpha & 1 & 0 & -T_\alpha & -C_\alpha^{-1} & 0 \\ T_\alpha & 0 & 1 & -T_\alpha & 0 & -C_\alpha^{-1} \end{bmatrix}, \quad (8)$$

where $T_\alpha = \tan(\alpha)$. Hence, the desired equations for the TCP displacement may be written as

$$\Delta p_x = \Delta \rho_x - \Delta L_x$$
$$\Delta p_y = T_\alpha \Delta \rho_x + \Delta \rho_y - T_\alpha \Delta L_x - C_\alpha^{-1} \Delta L_y \quad (9)$$
$$\Delta p_z = T_\alpha \Delta \rho_x + \Delta \rho_z - T_\alpha \Delta L_x - C_\alpha^{-1} \Delta L_z$$

It can be proved that similar results are valid for the "*YMax*" and "*ZMax*" postures (differing by the indices only), and also for the "*XMin*", "*YMin*", "*ZMin*" postures. In the latter case, the angle $\alpha$ should be computed as $\alpha = a\sin(\rho_{min}/L)$.

## 3 CALIBRATION METHOD

### 3.1 Measurement technique

To evaluate the leg/surface parallelism, we propose a single-sensor measurement technique. It is based on the fixed location of the measuring device for two distinct leg postures corresponding to the minimum/maximum values of the joint coordinates (Figure 4). Relevant calibration experiment consists of the following steps:

**Step 1**. Move the manipulator to the *"Zero"* posture; locate two gauges in the middle of the X-leg (parallel to the axes Y and Z); get their readings.

**Step 2**. Move the manipulator to the *"XMax"* and *"XMin"* postures, get the gauge readings, and compute differences.

**Step 3+**. Repeat steps 1, 2 for the Y- and Z-legs and compute corresponding differences.

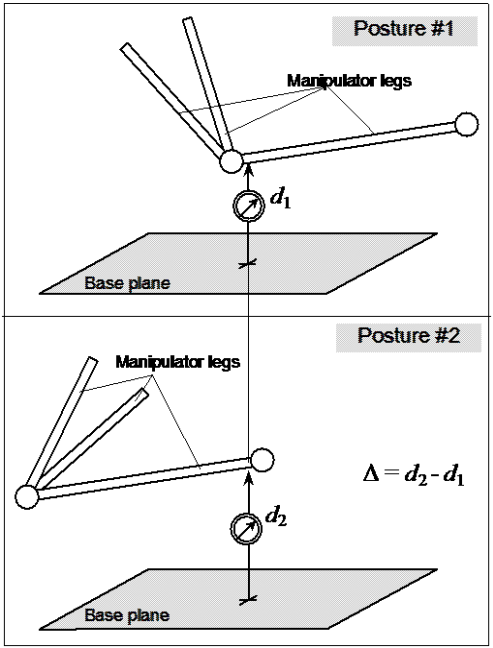

Figure 4: Measuring the leg/surface parallelism.

## 3.2 Calibration equations

The system of calibration equations can be derived in two steps. First, it is required to define the gauge initial locations that are assumed to be positioned at the leg middle at the "Zero" posture, i.e. at the points $(\mathbf{p}+\mathbf{r}_i)/2$, $i \in \{x, y, z\}$ where the vectors $\mathbf{r}_i$ define the prismatic joints centres: $\mathbf{r}_x = (L+\Delta\rho_x; 0; 0)$; $\mathbf{r}_y = (0; L+\Delta\rho_y; 0)$; $\mathbf{r}_z = (0; 0; L+\Delta\rho_z)$.

Hence, using the equation (7), the gauge initial locations can be expressed as

$$\mathbf{g}_x^0 = \left( \frac{L-\Delta L_x}{2} + \Delta\rho_x; \quad \frac{\Delta\rho_y - \Delta L_y}{2}; \quad \frac{\Delta\rho_z - \Delta L_z}{2} \right)$$

$$\mathbf{g}_y^0 = \left( \frac{\Delta\rho_x - \Delta L_x}{2}; \quad \frac{L-\Delta L_y}{2} + \Delta\rho_y; \quad \frac{\Delta\rho_z - \Delta L_z}{2} \right)$$

$$\mathbf{g}_z^0 = \left( \frac{\Delta\rho_x - \Delta L_x}{2}; \quad \frac{\Delta\rho_y - \Delta L_y}{2}; \quad \frac{L-\Delta L_z}{2} + \Delta\rho_z \right)$$

Afterwards, in the *"XMax"*, *"YMax"*, *"ZMax"* postures, the leg location is also defined by two points, namely, (i) the TCP, and (ii) the centre of the prismatic joint $\mathbf{r}_i$. For example, for the *"XMax"* posture, the TCP position is

$$\mathbf{p}_x^{max} = (LS_\alpha + \Delta\rho_x - \Delta L_x; \ *; \ *),$$

and the joint position is

$$\mathbf{r}_x^{max} = (L + LS_\alpha + \Delta\rho_x; \ 0; \ 0).$$

So, the leg is located along the line

$$\mathbf{s}_x(\mu) = \mu \cdot \mathbf{p}_x^{max} + (1-\mu) \cdot \mathbf{r}_x^{max},$$

where $\mu$ is a scalar parameter, $\mu \in [0, 1]$. Since the x-coordinate of the gauge is independent of the posture, the parameter $\mu$ may be obtained from the equation $[\mathbf{s}_x(\mu)]_x = [\mathbf{g}_x^0]_x$, which solution yields:

$$\mu = 0.5 + S_\alpha - S_\alpha \cdot \Delta L_x / L,$$

Hence, after some transformations, the deviations of the X-leg measurements (between the *"XMax"* and *"Zero"* postures) may be expressed as

$$\Delta y_x^+ = (0.5 + S_\alpha)T_\alpha \, \Delta\rho_x + S_\alpha \Delta\rho_y -$$
$$- (0.5 + S_\alpha)T_\alpha \Delta L_x - ((0.5 + S_\alpha)C_\alpha^{-1} - 0.5)\Delta L_y$$

$$\Delta z_x^+ = (0.5 + S_\alpha)T_\alpha \, \Delta\rho_x + S_\alpha \Delta\rho_z -$$
$$- (0.5 + S_\alpha)T_\alpha \Delta L_x - ((0.5 + S_\alpha)C_\alpha^{-1} - 0.5)\Delta L_z$$

Similar approach may be applied to the *"XMin"* posture, as well as to the corresponding postures for the Y- and Z-legs. This gives the system of twelve linear equations in six unknowns:

$$\begin{bmatrix} a_1 & b_1 & 0 & -c_1 & -b_1 & 0 \\ b_1 & a_1 & 0 & -b_1 & -c_1 & 0 \\ a_2 & b_2 & 0 & -c_2 & -b_2 & 0 \\ b_2 & a_2 & 0 & -b_2 & -c_2 & 0 \\ 0 & a_1 & b_1 & 0 & -c_1 & -b_1 \\ 0 & b_1 & a_1 & 0 & -b_1 & -c_1 \\ 0 & a_2 & b_2 & 0 & -c_2 & -b_2 \\ 0 & b_2 & a_2 & 0 & -b_2 & -c_2 \\ a_1 & 0 & b_1 & -c_1 & 0 & -b_1 \\ b_1 & 0 & a_1 & -b_1 & 0 & -c_1 \\ a_2 & 0 & b_2 & -c_2 & 0 & -b_2 \\ b_2 & 0 & a_2 & -b_2 & 0 & -c_2 \end{bmatrix} \begin{bmatrix} \Delta\rho_x \\ \Delta\rho_y \\ \Delta\rho_z \\ \Delta L_x \\ \Delta L_y \\ \Delta L_z \end{bmatrix} = \begin{bmatrix} \Delta x_y^+ \\ \Delta y_x^+ \\ \Delta x_y^- \\ \Delta y_x^- \\ \Delta y_z^+ \\ \Delta z_y^+ \\ \Delta y_z^- \\ \Delta z_y^- \\ \Delta z_x^+ \\ \Delta z_x^+ \\ \Delta x_z^- \\ \Delta z_x^- \end{bmatrix} \quad (10)$$

where

$a_i = S_{\alpha_i}$, $b_i = (0.5 + S_{\alpha_i})T_{\alpha_i}$, $c_i = (0.5 + S_{\alpha_i})C_{\alpha_i}^{-1} - 0.5$

and $\alpha_1 = \operatorname{asin}(\rho_{max}/L) > 0$, $\alpha_2 = \operatorname{asin}(\rho_{min}/L) < 0$.

This system can be solved using the pseudoinverse of Moore-Penrose, which ensures the minimum of the residual square sum.

## 4 EXPERIMENTAL RESULTS

The measuring system is composed of standard comparator indicators attached to the universal magnetic stands allowing fixing them on the manipulator bases. The indicators have resolution of 10 μm and are sequentially used for measuring the X-, Y-, and Z-leg parallelism while the manipulator moves between the *Max*, *Min* and *Zero* postures. For each measurement, the indicators are located on the mechanism base in such manner that a corresponding leg is admissible for the gauge contact for all intermediate postures (Figure 5).

For each leg, the measurements were repeated three times for the following sequence of motions: Zero → Max → Min → Zero→ …. Then, the results were averaged and used for the parameter identification (the repeatability of the measurements is about 0.02 mm).

To validate the developed calibration technique and the adopted modelling assumptions, there were carried out three experiments targeted to the following objectives: (#1) validation of modelling assumptions; (#2) collecting the experimental data for the parameter identification; and (#3) verification of the calibration results.

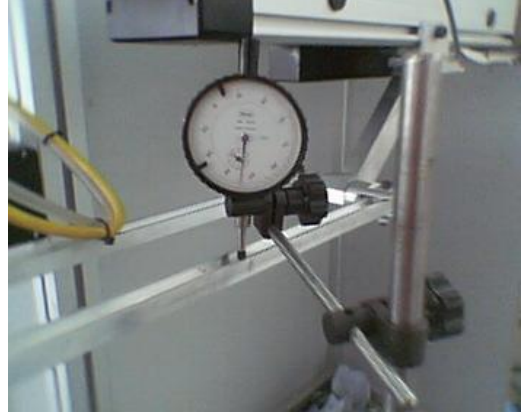

Figure 5: Experimental Setup.

Table 1: Calibration results.

| Parameters (mm) | | | | | | R.m.s. (mm) |
|---|---|---|---|---|---|---|
| $\Delta\rho_x$ | $\Delta\rho_y$ | $\Delta\rho_z$ | $\Delta L_x$ | $\Delta L_y$ | $\Delta L_z$ | |
| 4.66 | -5.36 | 1.46 | 5.20 | -5.96 | 3.16 | 0.12 |
| -0.48 | 0.49 | -1.67 | – | – | – | 0.14 |
| – | – | – | 0.50 | -0.52 | 1.69 | 0.14 |

The *first experiment* produced rather high parallelism deviation, which impels to conclude that the mechanism mechanics requires more careful tuning. Consequently, the location of the joint axes was adjusted mechanically to ensure the leg parallelism for the Zero posture.

The *second experiment* (after mechanical tuning) yielded lower deviations, twice better than for the first experiment. For these data, the developed calibration algorithm was applied for three sets of the model parameters: for the full set {Δρ, Δ*L*} and for the reduced sets {Δρ} and {Δ*L*}. As follows from the identification results (Table 1), the algorithms is able to identify simultaneously both the joint offsets and Δρ and the link lengths Δ*L*. However, both Δρ and Δ*L* (separately) demonstrate roughly the same influence on the residual reduction, from 0.32 mm to 0.14 mm, while the full set {Δρ, Δ*L*} gives further residual reduction to the 0.12 mm only. This motivates considering Δρ as the most essential parameters to be calibrated. Accordingly, the identified vales of the joint offsets were input into the control software.

The *third experiment* demonstrated good agreement with the expected results. In particular, the average deviation reduced down to 0.15 mm,

which corresponds to the measurement accuracy. On the other hand, further adjusting of the model to the new experimental data does not give the residual reduction.

Hence, the calibration results confirm validity of the proposed identification technique and its ability to tune the joint offsets and link lengths from observations of the leg parallelism. Other conclusion is related to the modelling assumption: for further accuracy improvement it is prudent to generalize the manipulator model by including parameters describing the orientation of the prismatic joint axes, i.e. relaxing assumption (iv) (see sub-section 2.2).

# 5 CONCLUSIONS

This paper proposes further developments for a novel calibration approach for parallel manipulators, which is based on observations of manipulator leg parallelism with respect to some predefined planes. This technique employs a simple and low-cost measuring system composed of standard comparator indicators, which are sequentially used for measuring the deviation of the relevant leg location while the manipulator moves the TCP along the Cartesian axes. From the measured differences, the calibration algorithm estimates the joint offsets and the link lengths that are treated as the most essential parameters to be tuned. The validity of the proposed approach and efficiency of the developed numerical algorithm were confirmed by the calibration experiments with the Orthoglide prototype, which allowed essential reduction of the residuals and corresponding improvement of the accuracy.

Future work will focus on the expanding the set of the identified model parameters, their identifiably analysis, and compensation of the non-geometric errors.

# REFERENCES

Chablat, D., Wenger, Ph., 2003. Architecture Optimization of a 3-DOF Parallel Mechanism for Machining Applications, the Orthoglide. *IEEE Transactions On Robotics and Automation*, Vol. 19 (3), pp. 403-410.

Daney, D., 2003. Kinematic Calibration of the Gough platform. *Robotica*, 21(6), pp. 677-690.

Huang, T., Chetwynd, D.G. Whitehouse, D.J., Wang, J., 2005. A general and novel approach for parameter identification of 6-dof parallel kinematic machines. *Mechanism and Machine Theory*, Vol. 40 (2), pp. 219-239.

Innocenti, C., 1995. Algorithms for kinematic calibration of fully-parallel manipulators. In: *Computational Kinematics*, Kluwer Academic Publishers, pp. 241-250.

Iurascu, C.C. Park, F.C., 2003. Geometric algorithm for kinematic calibration of robots containing closed loops. *ASME Journal of Mechanical Design*, Vol. 125(1), pp. 23-32.

Jeong J., Kang, D., Cho, Y.M., Kim, J., 2004. Kinematic calibration of redundantly actuated parallel mechanisms. *ASME Journal of Mechanical Design*, Vol. 126 (2), pp. 307-318.

Merlet, J.-P., 2000. *Parallel Robots*, Kluwer Academic Publishers, Dordrecht, 2000.

Pashkevich, A., Wenger, P., Chablat, D., 2005. Design Strategies for the Geometric Synthesis of Orthoglide-type Mechanisms. *Mechanism and Machine Theory*, Vol. 40 (8), pp. 907-930.

Pashkevich A., Chablat D., Wenger P., 2006. Kinematic Calibration of Orthoglide-Type Mechanisms. *Proceedings of IFAC Symposium on Information Control Problems in Manufacturing* (INCOM'2006), Saint Etienne, France, 17-19 May, 2006, p. 151 - 156

Renaud, P., Andreff, N., Pierrot, F., Martinet, P., 2004. Combining end-effector and legs observation for kinematic calibration of parallel mechanisms. *IEEE International Conference on Robotics and Automation* (ICRA'2004), New-Orleans, USA, pp. 4116-4121.

Renaud, P., Andreff, N., Martinet, P., Gogu, G., 2005. Kinematic calibration of parallel mechanisms: a novel approach using legs observation. *IEEE Transactions on Robotics* and Automation, 21 (4), pp. 529-538.

Tlusty, J., Ziegert, J.C., Ridgeway, S., 1999. Fundamental Comparison of the Use of Serial and Parallel Kinematics for Machine Tools. *CIRP Annals*, Vol. 48 (1), pp. 351-356.

Wang, J. Masory, O. 1993. On the accuracy of a Stewart platform - Part I: The effect of manufacturing tolerances. *IEEE International Conference on Robotics and Automation* (ICRA'93), Atlanta, Georgia, pp. 114–120.

Wenger, P., Gosselin, C., Chablat, D., 2001. Comparative study of parallel kinematic architectures for machining applications. In: Workshop on Computational Kinematics, Seoul, Korea, pp. 249-258.

Wenger, P., Gosselin, C. Maille, B., 1999. A comparative study of serial and parallel mechanism topologies for machine tools. In: *Proceedings of PKM'99*, Milan, Italy, pp. 23–32.